\def\year{2022}\relax
\documentclass[letterpaper]{article} 
\usepackage{aaai22}  
\usepackage{times}  
\usepackage{helvet}  
\usepackage{courier}  
\usepackage[hyphens]{url}  
\usepackage{graphicx} 
\usepackage{natbib}  
\usepackage{caption} 
\DeclareCaptionStyle{ruled}{labelfont=normalfont,labelsep=colon,strut=off} 
\usepackage{algorithm}
\usepackage{algorithmic}

\usepackage{amsmath,amsfonts,amsthm}

\usepackage{booktabs} 
\usepackage{tabularx}
\usepackage{multirow}
\usepackage{graphicx}
\usepackage{url}
\usepackage{array}
\usepackage{diagbox}
\usepackage[caption=false]{subfig}
\usepackage{adjustbox}
\usepackage{caption}
\newcolumntype{P}[1]{>{\centering\arraybackslash}p{#1}}
\newcommand{\subbcrfigsize}{0.244}  

\newcommand{\etal}{\textit{et al. }}

\usepackage{newfloat}
\usepackage{listings}
\floatstyle{ruled}
\newfloat{listing}{tb}{lst}{}
\floatname{listing}{Listing}
\title{Automatic Portrait Video Matting via Context Motion Network}

\title{Automatic Portrait Video Matting via Context Motion Network}
\author {
    Qiqi Hou\textsuperscript{\rm 1} \thanks{This project is done during Qiqi Hou's internship at Intel.}, \
    Charlie Wang \textsuperscript{\rm 2}
}
\affiliations {
    \textsuperscript{\rm 1} Portland State University
    \textsuperscript{\rm 2} Intel\\
    qiqi2@pdx.edu, charlie.l.wang@intel.com
}

\usepackage{bibentry}

\makeatletter
\g@addto@macro\@maketitle{
    \vspace*{-20pt}
    \tiny
    \centering
        \begin{tabular}{ccccc}
            \hspace{-2.5mm} \includegraphics[width=0.195\textwidth]{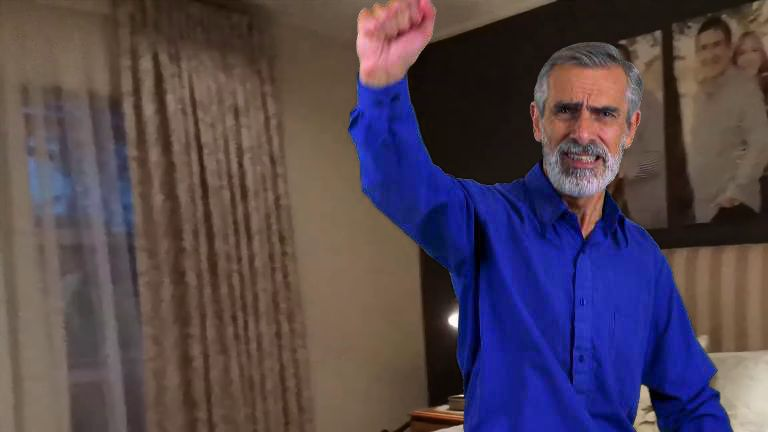} \hspace{-4mm} &
            \includegraphics[width=0.195\textwidth]{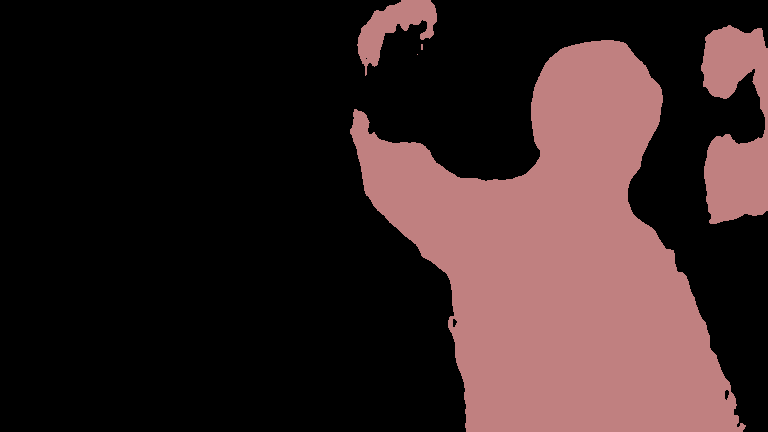}
            \hspace{-4mm} &
            \includegraphics[width=0.195\textwidth]{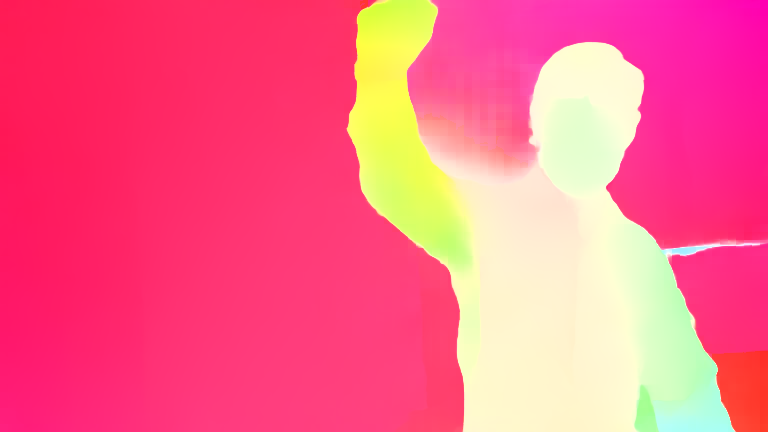}
            \hspace{-4mm} &
            \includegraphics[width=0.195\textwidth]{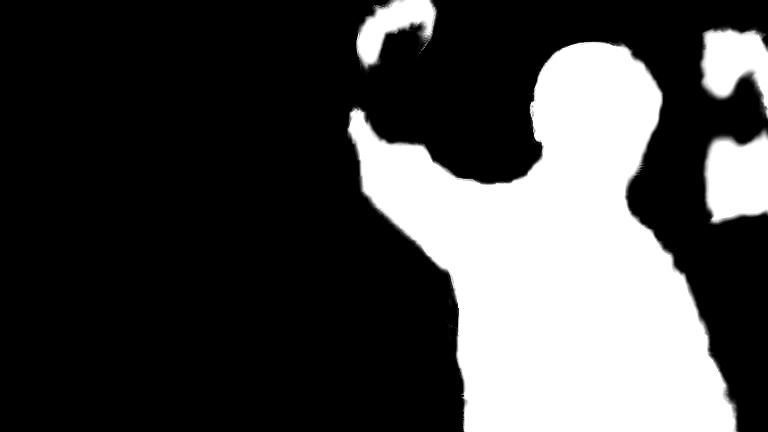}
            \hspace{-4mm} &
            \includegraphics[width=0.195\textwidth]{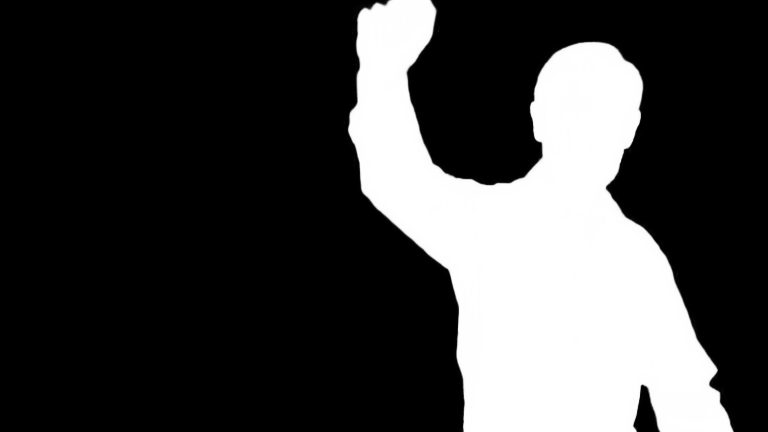} \\
            Frame & Segmentation~\cite{deeplabv3plus2018} & Optical flow & FBA Matting~\cite{forte2020f} & Ours
        \end{tabular}
   \captionof{figure}{
        Our automatic portrait video matting method does not require extra inputs. Most state-of-the-art matting methods rely on semantic segmentation methods to automatically generate the trimap. Their performance is compromised due to the lack of temporal information. Our method exploits semantic information as well as temporal information from optical flow and produces high-quality results. 
    }\vspace{0.05in}

\label{fig::figone} 
\vspace{-0.in}
}
\makeatother

\begin{document}

\maketitle

\begin{abstract}
Automatic portrait video matting is an under-constrained problem. Most state-of-the-art methods only exploit the semantic information and process each frame individually. Their performance is compromised due to the lack of temporal information between the frames. To solve this problem, we propose the context motion network to leverage semantic information and motion information. To capture the motion information, we estimate the optical flow and design a context-motion updating operator to integrate features between frames recurrently. Our experiments show that our network outperforms state-of-the-art matting methods significantly on the Video240K SD dataset.
\end{abstract}

\section{Introduction}
\label{sec:intro}
Portrait video matting aims to recover the foreground portraits from a portrait video. It has attracted extensive interest due to its wide applications in visual effects, video editing, and video meeting. It is an under-constrained problem.




To solve the matting problem, most existing methods rely on extra inputs to provide external information about the foreground or background areas. Trimap-based methods~\cite{xu2017deep,hou2019context,lu2019indices,li2020natural,forte2020f} require a trimap image as the extra input, where users specify known foreground and background areas, as well as an undefined area. These methods recover alpha matte values in the undefined area by propagating the alpha values from known area. These methods achieve promising performance. However, obtaining accurate trimaps often requires a vast amount of user effort. Alternatively, background-based methods~\cite{BMSengupta20,BGMv2} assume that the background is relatively static, and users can capture the background. The matting problem is reduced to estimate the matte values with a known background. These methods adopt an end-to-end network to estimate the alpha values directly and produce high-quality results. However, it is challenging to capture background for videos with moving backgrounds, which impedes their applications to more general scenarios. 

Our research is inspired by recent portrait image matting without any extra inputs~\cite{shen2016deep,zhu2017fast}. These methods incorporate the semantic information into the matting network to automatically estimate the alpha maps for portraits. For instance, Shen~\etal~\cite{shen2016deep} employs a FCN~\cite{long2015fully} as a part of the whole network to estimate the trimap. They estimate the matte values based on the trimap and achieve promising results. However, these methods are designed for images. Their performance on videos might be compromised due to the lack of temporal information. For instance, as shown in Figure~\ref{fig::figone}, the hand in the semantic segmentation is inaccurate, while the optical flow can provide useful temporal information for the hand. Furthermore, adopting these methods to each frame individually sometimes leads to undesirable temporal incoherence, such as flickering.





Our idea for portrait video matting is to make use of not only semantic information but also \textbf{motion} information from optical flow. There are several advantages of introducing motion information to the portrait video matting. First, the motions of foreground and background are typically different. It will be easier to find the edges between foreground and background. Second, if the foreground moves across the background, some portions of the background will be covered or revealed. We can construct a more complete background to matte against. 




We accordingly developed a context motion network dedicated to portrait video matting. Our network takes a sequence of frames as inputs and estimates the alpha maps and foregrounds for each frame. Specifically, our network contains four parts: the feature extraction network, the optical flow estimator, the context motion updating operator, and the upsampler. We first extract the context features for each frame and estimate the optical flow between consecutive frames. The optical flow and context features are fed into the context motion updating operator. To get the motion features, the context motion updating operator backwarps features from neighboring frames using optical flow and calculate the correlation between the current features and the backwarped features. Our operator also encodes the optical flow directly. We employ ConvGRUs to recurrently integrate features from consecutive frames. We then feed the resulting features to upsamplers to predict alpha maps and foreground. Our network is trained in an end-to-end manner. Our experiments show that our method outperforms the state-of-the-art trimap-based and background-based methods significantly.

This paper contributes to the research on automatic portrait video matting by leveraging semantic information and temporal information. First, we propose a novel context motion network. It is the first deep learning based method to utilize optical flow to capture the motion information in the automatic portrait video matting. Second, we design a context motion updating operator to fuse the context and temporal information and recurrently update features across consecutive frames. Third, the proposed method achieves state-of-the-art performances compared to the trimap-based and background-based methods.

\section{Related Work}
\label{sec:related}
Matting is a classic computer vision problem. For an image $\mathbf{I}$, it assumes that $\mathbf{I}$ is a linear composition of a foreground $\mathbf{F}$ and a background $\mathbf{B}$ based on an alpha map $\boldsymbol{\alpha} \in [0, 1]$,
\begin{equation}
    \mathbf{I} = \boldsymbol{\alpha} \mathbf{F} + (1 - \boldsymbol{\alpha}) \mathbf{B},
\end{equation}
where $\mathbf{F}, \mathbf{B}$, and $\boldsymbol{\alpha}$ are unknown. Matting aims to recover the foreground object, which is an under-constrained problem. In this part, we discuss the most relevant works on image matting and video matting. Due to the space limit, we refer readers to a comprehensive survey article Wang et.al.~\cite{wang2008image}.

\textbf{Image matting.} To solve the matting problem, most of the current methods rely on the extra inputs for external information. Trimap-based methods require a user-provided trimap or stokes which specify the foreground, background, and unknown areas. To recover the matte values in the unknown area, there are three categories of trimap-based matting methods: color sampling methods, propagation methods, and deep learning methods. Color sampling methods~\cite{chuang2001bayesian, wang2005iterative, he2011global} are based on the observation that neighboring pixels' alpha values are often similar if their colors are similar.
However, the performance of color sampling methods declines sharply when the color distributions of foreground and background have a large overlap. Propagation methods~\cite{aksoy2017designing, chen2013knn, chen2013image, he2010fast, lee2011nonlocal, levin2008closed, levin2008spectral,sun2004poisson, levin2008closed, aksoy2017designing, chen2013knn} propagate alpha values from known to unknown pixels through various affinities between neighboring pixels. 
However, it is challenging to obtain accurate affinities because of color ambiguity. 
Deep learning methods use a CNN and estimate the alpha map from the source images ~\cite{shen2016deep,xu2017deep,cho2016natural, cho2019deep,hou2019context,lu2019indices,tang2019learning,li2020natural} or finetune the results from traditional methods~\cite{cho2016natural,cho2019deep}. Instead of using hand-crafted features, these methods learned the features which are more robust to the challenging scenarios. These methods achieved promising results. However, trimap-based methods require an amount of user effort in creating the trimap. The extensive requirements of accurate trimaps pose a challenge, especially when running these methods on videos. 

Alternatively, background-based methods~\cite{BMSengupta20,BGMv2} assume that the background image is relatively static, and users can capture the background image. The matting problem is reduced to estimate the foreground objects with a known background. These methods take the image as well as the background image as inputs and estimate the alpha mattes with an end-to-end trained network, which have shown particularly effective and high efficiency when the background is relatively static. However, their performance drops sharply when it has a large mismatch between the user-captured background and the real background. Furthermore, for the videos with a moving background, it is difficult to capture the accurate background. 

The semantic-based portrait matting methods~\cite{shen2016deep,zhu2017fast} leverage the semantic information to estimate the matte values for the portrait matting without any user interaction. For instance, Shen~\etal accordingly proposed an end-to-end trained network~\cite{shen2016deep}. It first estimated the trimap. Then they designed a matting layer to estimate the matte values based on the trimap. Their method achieved promising results for images. Compared to these methods, our method improves the accuracy of automatic matting with temporal information. 

Recently, some methods~\cite{li2020hierarchical,qiao2020attention} estimate the matte values for any objects without any extra inputs. However, the generalization capability of these methods is compromised.

\textbf{Video matting.} Most traditional video matting methods solve the video matting problem in two steps~\cite{chuang2002video,apostoloff2004bayesian,lee2010temporally,bai2011towards}. In the first step, these methods employ semantic segmentation methods to estimate a binary segmentation image of each frame independently. Based on the binary segmentation, these methods generate trimaps for each frame by performing spatio-temporal optimizations. In the second step, these methods apply image matting algorithms to estimate the matte values for each frame using trimaps generated in the first step. While these methods have achieved promising results for some scenes, the quality and stability of predictions from these methods decline when they are applied to challenging scenarios because these methods heavily rely on hand-crafted features.  Our method addresses this challenging problem by learning the features with an end-to-end trained network.

Recently, there are deep-learning based methods for video matting~\cite{ke2020green,zhang2021attention,sun2021deep}. For instance, Ke~\etal first estimates the alpha maps for each frame and employs a self-supervised strategy to preserve the temporal coherence~\cite{ke2020green}. Zhang~\etal employs an attention-based temporal aggregation module to recurrently predicts the alpha maps for videos~\cite{zhang2021attention}. Sun~\etal fuses the features from multiple frames to estimate the alpha maps for the target frame~\cite{sun2021deep}. Compared to these methods, our method use optical flow to capture the motion information. 

There is also a concurrent work by by Lin~\etal~\cite{rvm} that also applies temporal information for video matting. But it targets efficient video matting, while our method focuses on the high quality video matting. To facilitate comparisons between these two methods, we both compare to a previous baseline method BGMV2~\cite{BGMv2}.

\section{Method}
\label{sec:method}
\begin{figure*}[h]
\centering
\footnotesize
  \includegraphics[width=0.96\textwidth]{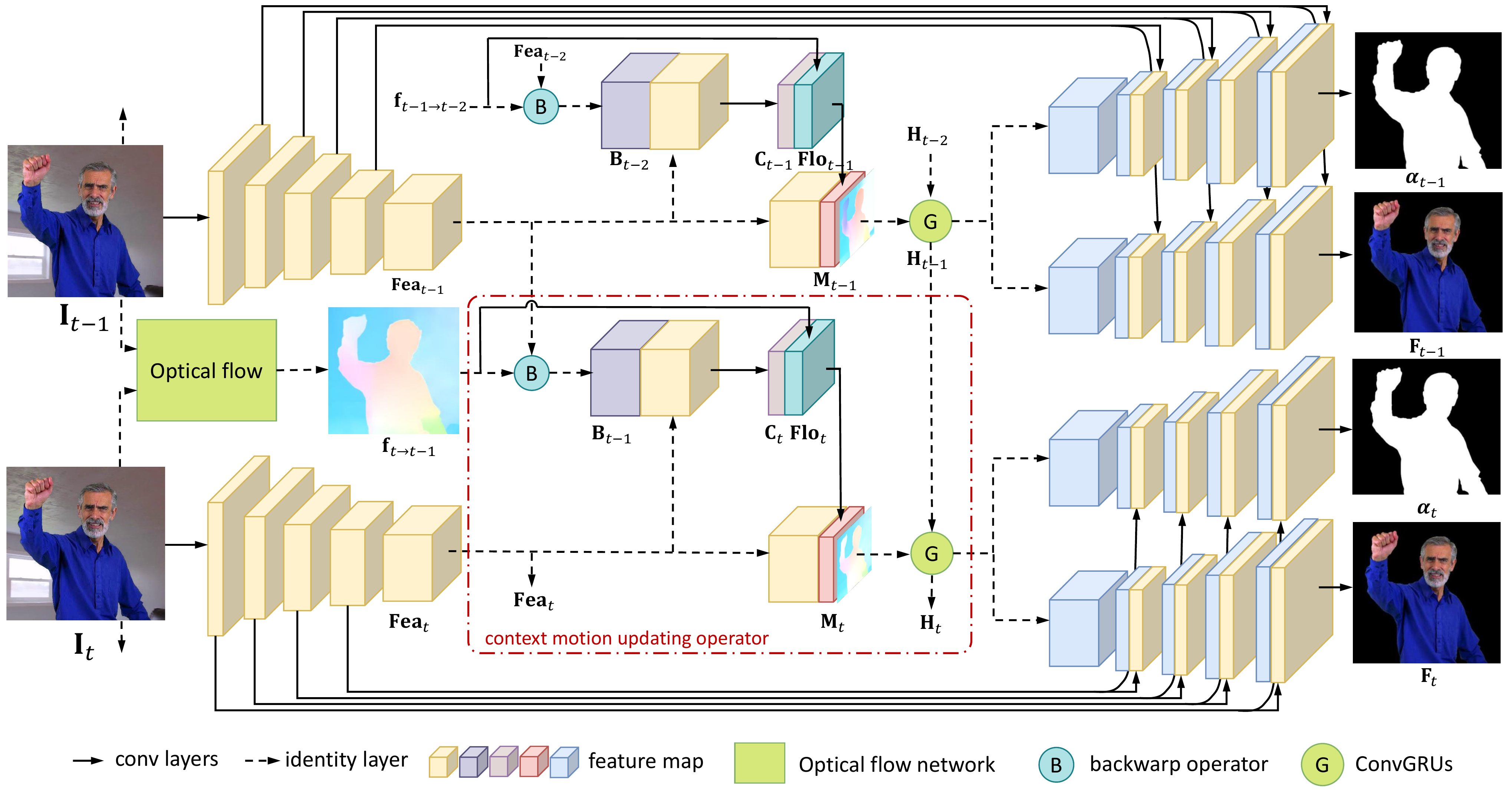}\vspace{-0.15in}
  \caption{The architecture of our network. Our network takes a sequence of frames as inputs and estimate the alpha maps and foreground image for each frame.}
  \label{fig:network}\vspace{-0.2in}
\end{figure*}

Our method takes a sequence of portrait frames $\{\mathbf{I}_t\}_{t=1}^N$ as inputs and aims to estimate their corresponding alpha maps $\{ \boldsymbol{\alpha}_t\}_{t=1}^N$ as well as their foregrounds  $\{\mathbf{F}_t\}_{t=1}^N$, where $N$ indicates the number of frames. Our network leverages not only semantic information but also \textbf{temporal} information. 

As shown in Figure~\ref{fig:network}, we develop a new deep network architecture for automatic portrait video matting. It recurrently estimates the alpha maps and foreground images. Our network can be distilled into four parts: (1) feature extraction, (2) optical flow estimation, (3) context-motion updating operator, and (4) upsampler. All parts are differentiable, and we train the network in an end-to-end manner. Given an input frame $\mathbf{I}_t$, our network first extracts the context features $\mathbf{Fea}_t$ from an encoder pretrained on the image classification task.  We capture the motion information by estimating the backward optical flow $\mathbf{f}_{t \rightarrow t-1}$ from the current frame $\mathbf{I}_t$ to the previous frame $\mathbf{I}_{t-1}$ by the PWC-Net~\cite{sun2018pwc}. To fuse the context information and the motion information, we develop a context-motion updating operator, which integrates context features, optical flow as well as the features from the previous frame. Finally, we upscale the resulting features $\mathbf{H}_t$ and predict the alpha maps $\boldsymbol{\alpha}_t$ and the foreground $\mathbf{F}_t$. Below we describe the network in detail.

\textbf{Feature extraction.} We extract the features from the input images using a convolutional network. The feature encoder network is applied to all the input frames. It maps the input images to feature maps at a lower resolution as follows
\begin{equation}
    \mathbf{Fea}_t = g_{encoder}(\mathbf{I}_t),
\end{equation}
where $g_{encoder}$ indicates the encoder network. We adopt the modified ResNet50~\cite{he2016deep} pretrained on the ImageNet dataset. Our encoder outputs features at $1/16$ resolution of the source frame as we empirically find that the large receptive field is critical for the semantic information. Aside from the context features, we also extract the multi-scale intermediate features from the encoder to capture the fine structures. We use $3\times3$ convolutional layers to map these intermediate feature maps to 32 channels. The weights of the feature extraction network are shared for all the input frames.

\textbf{Optical flow estimation.} We calculate the optical flow between the current frame $\mathbf{I}_t$ and the previous frame $\mathbf{I}_{t-1}$, which indicates the per-pixel motion between these two frames. Inspired by the success of state-of-the-art optical flow networks~\cite{sun2018pwc,zhao2020maskflownet,teed2020raft}, we leverage the PWC-Net~\cite{sun2018pwc} as a sub-network in our network to estimate the optical flow. It takes two consecutive video frames as input and output the backward optical flow as follows,
\begin{equation}
    \mathbf{f}_{t \rightarrow t - 1} = g_{flow}(\mathbf{I}_t, \mathbf{I}_{t-1}),
\end{equation}
where $g_{flow}$ indicates the optical flow network. $\mathbf{f}_{t \rightarrow t - 1} \in \mathbb{R}^{H \times W \times 2}$ indicates the backward optical flow. We initialize the the optical flow subnetwork using weights pretrained on the Sintel dataset~\cite{Butler:ECCV:2012}. During our training, we train the whole network in an end-to-end manner. 

\textbf{Context-motion updating operator.} Context-motion updating operator is designed to fuse semantic information and temporal information. The context-motion updating operator is recurrent. The features from the previous frames are utilized in the context-motion updating operator to provide temporal information.  

 
Given the frame feature $\mathbf{Fea}_t$, $\mathbf{Fea}_{t-1}$ as well as the backward optical flow $\mathbf{f}_{t \rightarrow t - 1}$, our goal is to capture the motion information. Inspired by the design of PWCNet~\cite{sun2018pwc}, we backwarp the features $\mathbf{Fea}_{t-1}$ from previous frame to the current frame by
\begin{equation}
    \mathbf{B}_{t - 1} = g_{backwarp}(\mathbf{Fea}_{t - 1}, \mathbf{f}_{t \rightarrow t - 1}),
\end{equation}
where $g_{backwarp}(\cdot)$ indicates the operator of backwarping. $\mathbf{B}_{t - 1}$ indicates the backwarped features. We calculate the correlation between the current feature $\mathbf{Fea}_t$ and the backwarped feature $\mathbf{B}_{t - 1}$ 
\begin{equation}
    \mathbf{C}_t = g_{corr}([\mathbf{Fea}_t, \mathbf{B}_{t-1}]),
\end{equation}
where $[\cdot]$ indicates the concenating operator. $g_{corr}$ indicates the operator to calculate the correlation. Our network adopts two $3 \times 3$ convolutional layers with 32 channels. 

Our network also encodes the optical flow to capture the motion information from optical flow directly. We first bilinear downsample the optical flow size to the size of $\mathbf{Fea}_t$. The optical flow is encoded as 
\begin{equation}
    \mathbf{Flo}_t = g_{flow\_encode}(\mathbf{f}_{t \rightarrow t - 1}),
\end{equation}
where $g_{flow\_encode}$ indicates the operator for optical flow encoding. Our network adopts two $7 \times 7$ convolutional layers with 32 channels. 

We fuse the correlation $\mathbf{C}_t$ and optical flow feature $\mathbf{Flo}_t$ to get the motion features
\begin{equation}
    \mathbf{M}_t = g_{motion}([\mathbf{C}_t, \mathbf{Flo}_t ]),
\end{equation}
where $g_{flow\_encode}(\cdot)$ indicates the operator for motion encoding.  Our network adopts a $3 \times 3$ convolutional layer with 62 channels. 

By concatenating the semantic feature $\mathbf{Fea}_t$, motion feature $\mathbf{M}_t$, as well as the optical flow $\mathbf{f}_{t \rightarrow t - 1 }$, our method gets the fused feature $\mathbf{Fus}_t$,
\begin{equation}
    \mathbf{Fus}_t = [\mathbf{Fea}_t, \mathbf{M}_t, \mathbf{f}_{t \rightarrow t - 1 }].
\end{equation}


Inspired by RAFT~\cite{teed2020raft}, our network adopts separable ConvGRUs to leverage the features from the previous frame. It replaces the fully connected layers in the GRU with separable convolutions. It can be represented as 
\begin{equation}
    \mathbf{z}_t = g_z([\mathbf{H}_{t-1}, \mathbf{Fus}_t]),
\end{equation}
\begin{equation}
    \mathbf{r}_t = g_r([\mathbf{H}_{t-1}, \mathbf{Fus}_t])
\end{equation}
\begin{equation}
    \widetilde{\mathbf{H}_t} = g_h([\mathbf{r}_t \cdot \mathbf{H}_{t - 1}, \mathbf{Fus}_t])
\end{equation}
\begin{equation}
    \mathbf{H}_t = (1 - \mathbf{z}_t) \cdot \mathbf{H}_{t - 1} + \mathbf{z}_t \cdot \widetilde{\mathbf{H}_t},
\end{equation}
where $g_z(\cdot)$,  $g_r(\cdot)$, and $g_h(\cdot)$ indicate the gated activation units. Specifically, $g_z(\cdot)$ and $g_r(\cdot)$ employs $sigmoid(\cdot)$ as their activation function. $g_h(\cdot)$ uses $tanh(\cdot)$. Our network adopts two ConvGRUs. In the first ConvGRU, $g_z(\cdot)$,  $g_r(\cdot)$ and $g_h(\cdot)$ are one $1\times5$ convolutional layer. And in the second ConvGRU, they are $5\times1$ convolutional layers. It can increase the receptive field while keeps the model size small.

\textbf{Upsampler.} As shown in Figure~\ref{fig:network}, our network contains two decoders: one for the alpha maps and the other for the foreground. They have the identity network architecture. We first bilinearly upsample the features by a factor of 2. Our network concatenates the upsampled features with the intermediate features from the encoder. Our network fuses them by two $3\times3$ convolutional layers with 32 channels. We repeat the process until the resolution of upsampled features is the same as the input frame. The last layer of the alpha decoder has two convolutional layers. We use the activation function $sigmoid()$ for the second convolutional layer. The last layer of the foreground decoder also has two convolutional layers. The second convolutional layer direct outputs the foreground image without activation function.

\textbf{Loss functions.} For each frame of the input video, we compute the loss over the alpha maps and the foreground images. For the alpha maps, we use the standard $\ell_1$ to measure the difference between the predicted alpha maps $\boldsymbol{\alpha}_t$ and the ground truth $\widehat{\boldsymbol{\alpha}}_t$ as follows
\begin{equation}
    \mathcal{L}_{1, t}^{\alpha} = \Vert \boldsymbol{\alpha}_t - \widehat{\boldsymbol{\alpha}}_t \Vert_1 .
\end{equation}

Following Context Matting~\cite{hou2019context}, we also employ the Laplacian loss for the alpha maps. Instead of measuring the difference from the image directly, it measures the differences of two Laplacian pyramid as follows,
\begin{equation}
    \mathcal{L}_{lap, t}^{\alpha} = \sum_{i = 1}^5 2 ^{i - 1} \Vert L^{i} (\boldsymbol{\alpha}_t) -  L^{i} (\widehat{\boldsymbol{\alpha}}_t) \Vert_1, 
\end{equation}
where $L^i(\cdot)$ indicates the $i^{th}$ level of the Laplacian pyramid. 

For the foreground image, we also employ the standard $\ell_1$ to measure the difference between the predicted foreground image $\mathbf{F}_t$ and $\widehat{\mathbf{F}}_t$. We only calculate the loss where the foreground is visible
\begin{equation}
    \mathcal{L}_{1, t}^{fg} = \Vert \mathbb{I} (\widehat{\boldsymbol{\alpha}}_t > 0) \cdot (\mathbf{F}_t - \widehat{\mathbf{F}}_t) \Vert_1,
\end{equation}
where $\mathbb{I}(\cdot)$ indicates a binary function whose value is 1 if the condition is true and 0 otherwise.

We get the total loss as
\begin{equation}
    \mathcal{L}_{total} = \sum_{t = 1}^{N}\mathcal{L}_{1, t}^{\alpha} + \mathcal{L}_{lap, t}^{\alpha} + 0.1\mathcal{L}_{1, t}^{fg},
\end{equation}
where $N$ indicates the length of the input video.

\textbf{Training.} We use PyTorch to train our neural network. The feature extraction network is initialized with the official weights pretrained on the ImageNet dataset. And the optical flow network is initialized with the weights pretrained on the Sintel dataset. The other modules are initialized with random weights. We adopt AdamW~\cite{loshchilov2017decoupled} as our optimizer. Our initial learning rate is set to $10^{-4}$. We use the ``OneCycleLR'' learning rate policy ~\cite{smith2019super} in training, which we empirically find it converges faster than other policies.  Furthermore, we clip gradients in the range $[-1, 1]$. 

We train our network for 300k iterations on a single NVIDIA Titan RTX GPU. In the first 150k iterations, we freeze the parameters of the optical flow network. We empirically find that freezing the parameters of the optical flow network in the early stages can make the training process more stable. Our batchsize is set to 4. In the following 150k iterations, we unfreeze the parameters of the optical flow network and train the network in an end-to-end manner. We set the batchsize to 2 due to the GPU memory limitation.  We set the length of frames as 3 for training. Furthermore, our network freezes the parameters of batch norm layers during the whole training process.

\begin{figure}
    \centering
    \includegraphics[width=0.48\textwidth]{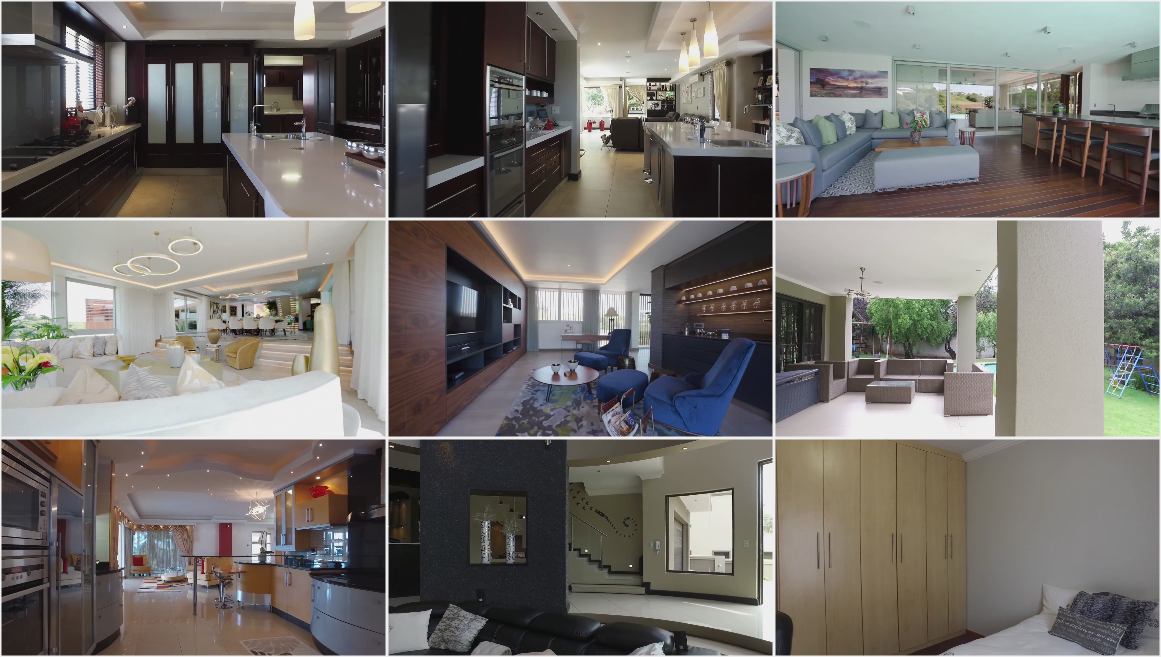}
    \caption{Examples from our background video dataset. It contains various indoor scenarios.}
    \label{fig:bg} \vspace{-0.2in}
\end{figure}

\textbf{Dataset}. We train our network using the video portrait matting dataset, Video240K SD, shared by ~\cite{BGMv2}. This dataset contains 484 training videos with the corresponding alpha maps and the foreground images, including 479 videos for training and 5 videos for testing. The resolution of videos is $768\times432$. The Video240K SD dataset only releases the background images, while our method is designed for the videos. We collected 82 indoor videos from YouTube as the background videos. These videos are under Common Creative Licenses. The resolution of these videos is higher than or equal to 1080p. Most videos last more than 10 minutes, and the frame rate is higher than or equal to 24fps. We randomly extracted frame sequences from these videos. Each frame sequence contains 50 consecutive frames. We also downsampled the frames to 720p. We manually removed the sequences with cut shots, visual effects, and extremely fast motion. We also manually removed the sequences which contain humans. This culling process reduced the total number of sequences to 5636, consisting of 281,800 individual video frames with a typical resolution of $1080\times720$. We split them into two parts: 5586 from 68 videos as the training background and 50 from 14 videos as the testing background. There is no overlap scene among them. As shown in Figure~\ref{fig:bg}, our dataset covers various indoor scenarios. For training, we created the training samples by compositing the foreground video onto a randomly selected background sequences. For the testing, we composite each testing foreground video onto 10 randomly selected background videos. There are 50 testing videos in total. 

\textbf{Data augmentation}. As discussed in ~\cite{hou2019context}, the foreground videos and background videos often contain different artifacts. Directly training the network on the composited videos without augmentation might compromise the generalization capability of the trained network. Following ~\cite{hou2019context}, we leveraged the JPEG noise to the composited videos. We keep the quality of the composited videos in the range of [70\%, 100\%]. Aside from the augmentations of images, we also introduced the H264 noise to our composited videos, which is commonly used in video compression. Our method also uses standard augmentations. Specifically, we randomly crop the composited videos to patches of size $384 \times 384$, as the large cropping size is critical for semantic information. We randomly flip the images horizontally. We employ the color jitter for the foreground video to increase the color diversity. We set the brightness in a range of [0.85, 1.15], the contrast in the range of [0.85, 1.15], the hue in the range of [0.9, 1.1], the saturation in the range of [0.7, 1.3]. We also employ the gamma transformation for the alpha maps. We set the gamma value in the range of [0.2, 2].

\section{Experiment}
\label{sec:exp}
\begin{table}
    \centering
    \scriptsize
    \begin{tabular}{P{1.4cm}P{0.62cm}P{0.62cm}P{0.62cm}P{0.62cm}P{0.62cm}P{0.9cm}}
            \toprule
            \multirow{2}[2]{*}{Method} & Extra & \multicolumn{4}{c}{Alpha} & Foreground  
            \\ \cmidrule(lr){3-6} 
             &input &SAD &MSE &Grad &Conn &MSE
            \\ \midrule
            \multirow{2}[0]{*}{Context Matting} & $\mathbf{T}_{seg}$ & 5.39 & 13.38 & 18.09 & 5.34 & 5.17 \\
                                            & $\mathbf{T}_{gt}$   & 1.41 & 1.78  & 4.24  & 1.31 & 4.26 \\
            \multirow{2}[0]{*}{Index Matting} & $\mathbf{T}_{seg}$ & 4.32 & 10.32 & 13.03 & 4.22 & - \\
                                              & $\mathbf{T}_{gt}$   & 1.47 & 1.57  & 2.74  & 1.28 & - \\
            \multirow{2}[0]{*}{GCA Matting}   & $\mathbf{T}_{seg}$ & 4.33 & 10.91 & 14.14 & 4.24 & - \\
                                              & $\mathbf{T}_{gt}$   &1.49  & 2.32  & 3.29  & 1.37 & - \\
            \multirow{2}[0]{*}{FBA Matting} & $\mathbf{T}_{seg}$ & 4.17 & 10.18 & 12.36 & 4.10 & 2.50 \\
                                            & $\mathbf{T}_{gt}$   & 1.02 & 0.87  & 1.84  & 0.91 & 2.25 \\
            \multirow{2}[0]{*}{BGMV2}  & $\mathbf{B}_{first}$ & 83.56 &247.03 &16.79 &- &3.79  \\
                                              & $\mathbf{B}_{gt}$    &1.07 & 1.41 & 2.47 & 0.94 & 4.01\\  \midrule
            Ours & - & \textbf{0.44} & \textbf{0.27} & \textbf{0.38} & \textbf{0.26} & \textbf{0.88}  
            \\ \bottomrule
    \end{tabular}\vspace{-0.1in}
        \caption{Comparison on the Video240K SD dataset. $\mathbf{T}_{seg}$ indicates the trimap generated from the semantic segmentation method~\cite{deeplabv3plus2018}. $\mathbf{T}_{gt}$ indicates the trimap generated from the ground truth alpha maps. $\mathbf{B}_{first}$ indicates the background image from the first frame. $\mathbf{B}_{gt}$ indicates the ground truth background image for each frame in the video.  \label{table:comp_video240}} \vspace{-0.3in}
\end{table}

\begin{figure*}[h]
    \newlength\indentspace
    \setlength{\indentspace}{-3.12mm}
    \centering
    \hspace{-3.91mm} 
    \begin{adjustbox}{valign=t}
        \scriptsize
            \begin{tabular}{cccc}
                \includegraphics[width=\subbcrfigsize\textwidth]{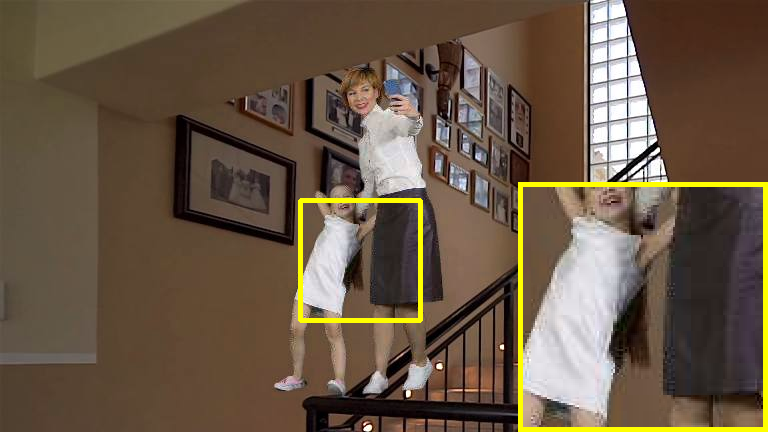}\hspace{\indentspace} &
                \includegraphics[width=\subbcrfigsize\textwidth]{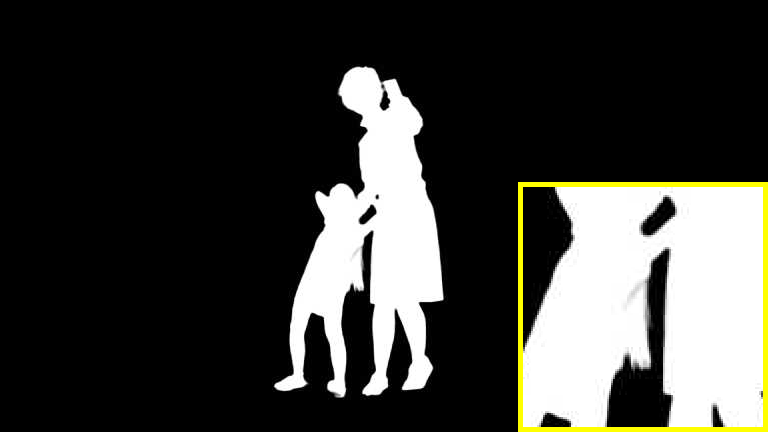}\hspace{\indentspace} &
                \includegraphics[width=\subbcrfigsize\textwidth]{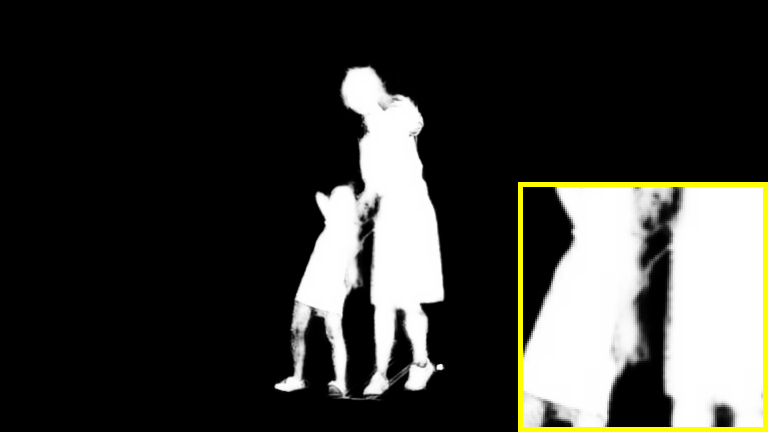}\hspace{\indentspace} &
                \includegraphics[width=\subbcrfigsize\textwidth]{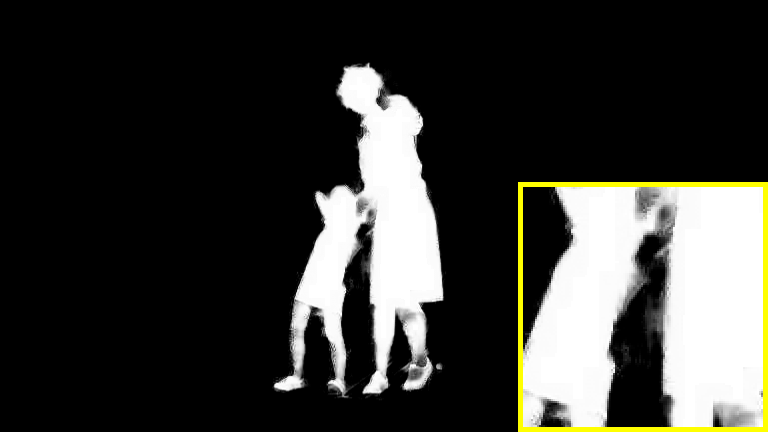}
                \\
                Frame \hspace{\indentspace} &
                Ground Truth Alpha\hspace{\indentspace} &
                Context Matting~\cite{hou2019context}\hspace{\indentspace} &
                Index Matting~\cite{lu2019indices}
                \\
                \includegraphics[width=\subbcrfigsize\textwidth]{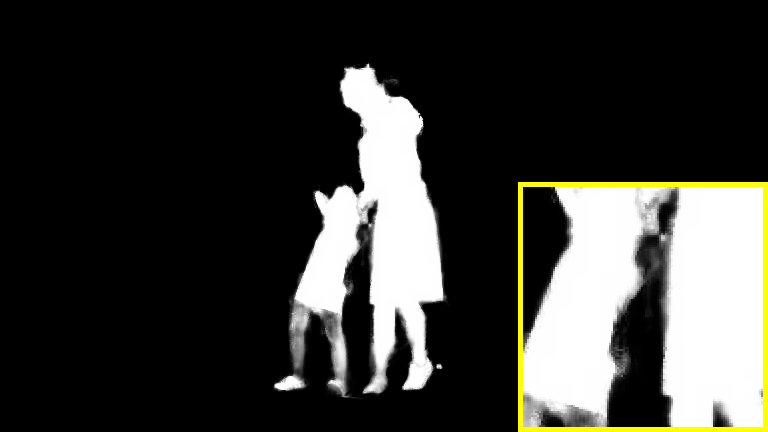}\hspace{\indentspace} &
                \includegraphics[width=\subbcrfigsize\textwidth]{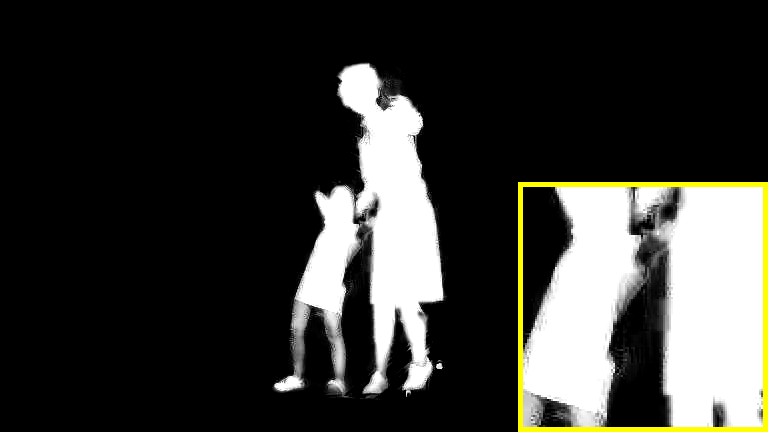}\hspace{\indentspace} &
                \includegraphics[width=\subbcrfigsize\textwidth]{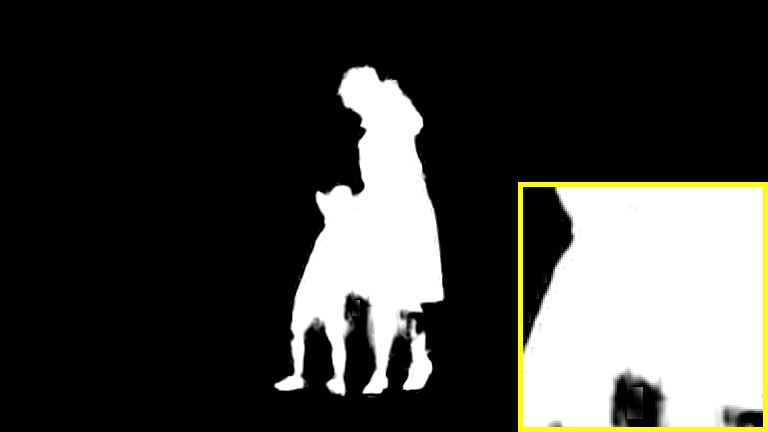}\hspace{\indentspace} &
                \includegraphics[width=\subbcrfigsize\textwidth]{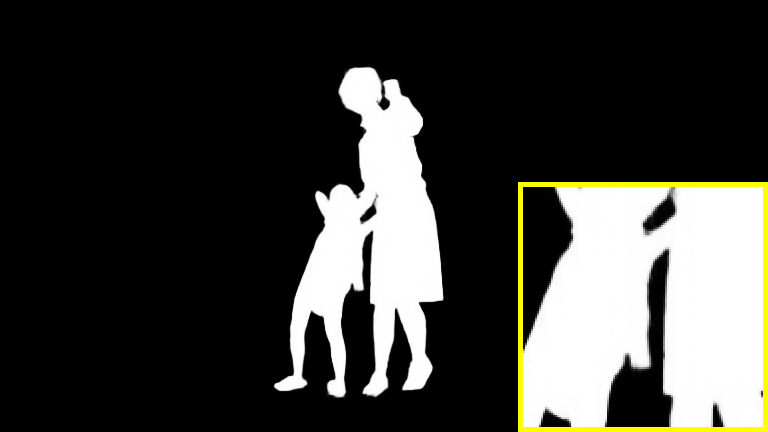}
                \\
                GCA Matting~\cite{li2020natural} \hspace{\indentspace} &
                FBA Matting~\cite{forte2020f}  \hspace{\indentspace} &
                BGMV2~\cite{BGMv2} \hspace{\indentspace} &
                Ours 
            \end{tabular}
        \end{adjustbox}
        \\
    \hspace{-3.91mm}
        \begin{adjustbox}{valign=t}
        \scriptsize
            \begin{tabular}{cccc}
                \includegraphics[width=\subbcrfigsize\textwidth]{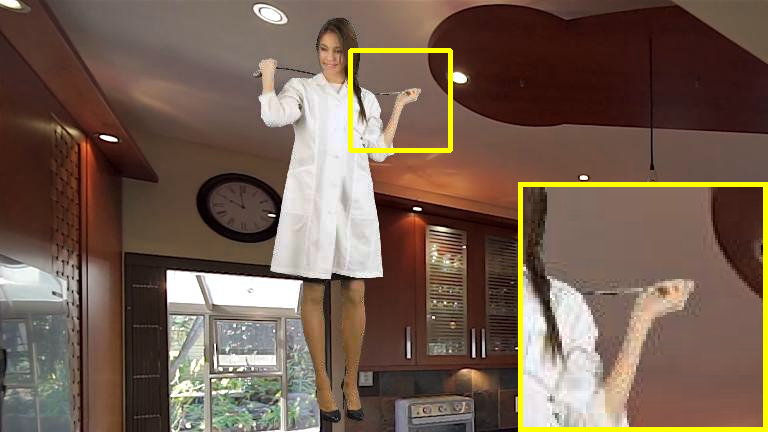}\hspace{\indentspace} &
                \includegraphics[width=\subbcrfigsize\textwidth]{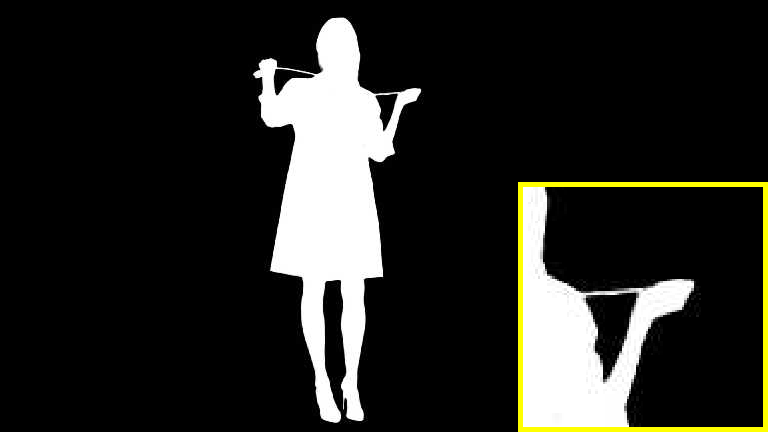}\hspace{\indentspace} &
                \includegraphics[width=\subbcrfigsize\textwidth]{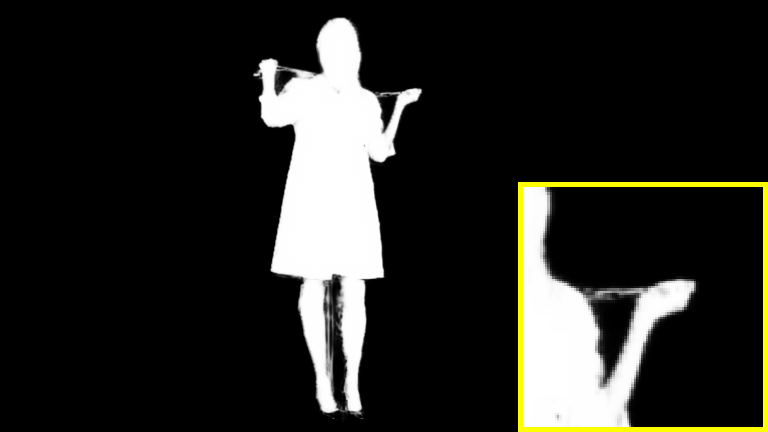}\hspace{\indentspace} &
                \includegraphics[width=\subbcrfigsize\textwidth]{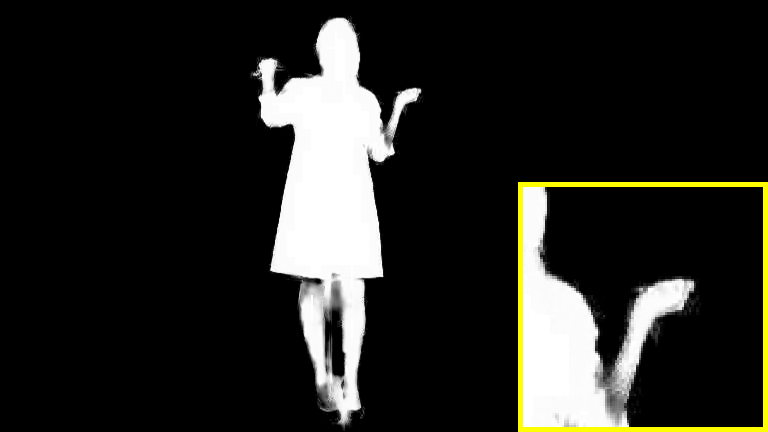}
                \\
                Frame \hspace{\indentspace} &
                Ground Truth Alpha\hspace{\indentspace} &
                Context Matting~\cite{hou2019context}\hspace{\indentspace} &
                Index Matting~\cite{lu2019indices}
                \\
                \includegraphics[width=\subbcrfigsize\textwidth]{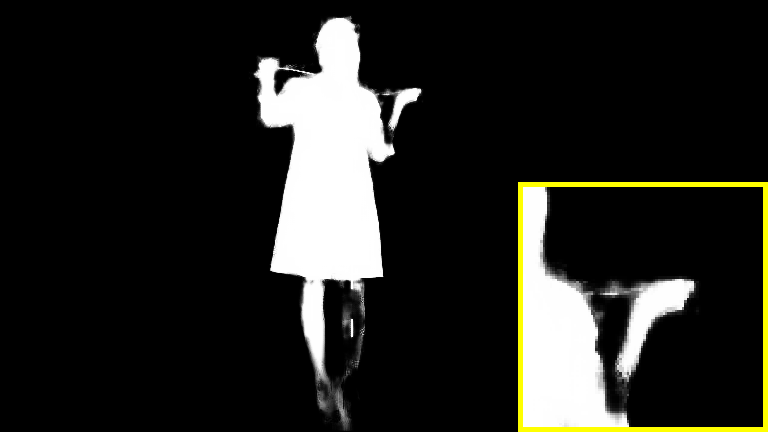}\hspace{\indentspace} &
                \includegraphics[width=\subbcrfigsize\textwidth]{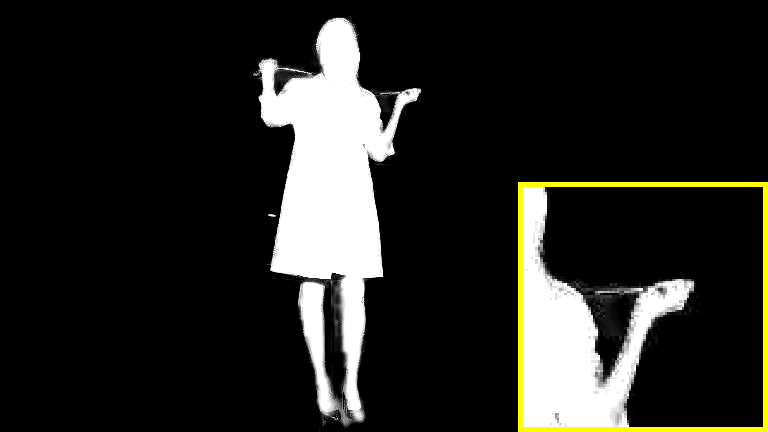}\hspace{\indentspace} &
                \includegraphics[width=\subbcrfigsize\textwidth]{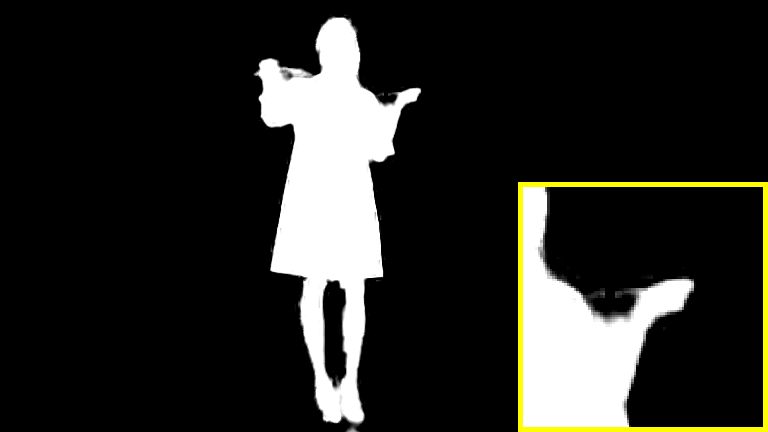}\hspace{\indentspace} &
                \includegraphics[width=\subbcrfigsize\textwidth]{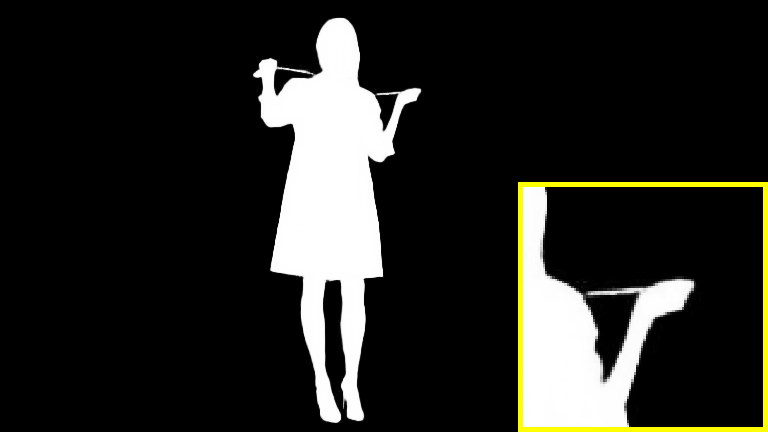}
                \\
                GCA Matting~\cite{li2020natural} \hspace{\indentspace} &
                FBA Matting~\cite{forte2020f}  \hspace{\indentspace} &
                BGMV2~\cite{BGMv2} \hspace{\indentspace} &
                Ours 
            \end{tabular}
        \end{adjustbox}
        \\
    \vspace{-0.1in}
    \caption{
        Visual comparison on the Video240K SD dataset. Our method generates more plausible results. We adopt the trimaps $\mathbf{T}_{gt}$ generated from the ground truth alpha maps for the trimap-based methods, including Index Matting~\cite{lu2019indices}, Context Matting ~\cite{hou2019context}, GCA Matting~\cite{li2020natural} and FBA Matting~\cite{forte2020f}. For the BGMV2~\cite{BGMv2}, we adopt the ground truth background $\mathbf{B}_{gt}$ for it. \label{fig:comp_video240k}
    } \vspace{-0.25in}
\end{figure*}

We conduct our experiments on the Video240K SD dataset. To compare with the trimap-based methods, we generate two versions of trimaps. The first version $\mathbf{T}_{gt}$ follows the Adobe Matting Dataset~\cite{xu2017deep}, which is generated by dilating the ground truth alpha mattes. For each video, we randomly select the size of dilation in the range of $[5, 15]$. Since manually creating the trimaps will take a lot of user effort, some users automatically create the trimaps by dilating the results from the semantic segmentation networks. As a consequence, we generate the second version $\mathbf{T}_{seg}$ by dilating the semantic segmentation using the Deeplab V3+ ~\cite{deeplabv3plus2018}. We set the size of dilation as 10 pixels. 

\textbf{Evaluation metrics}. For the alpha maps, we adopt various metrics, including SAD, MSE, Gradient (Grad), and Connectivity (Conn)~\cite{xu2017deep}. For the foreground videos, we evaluate the MSE of the foreground video. In our experiments, we employ the official evaluation codes from the authors of the Video240K SD dataset~\cite{BGMv2}. We scale MSE, Grad, Conn by $10^3$, $10^{-3}$ and $10^{-3}$, respectively. Please note that the evaluation codes only measure results of unknown areas on the trimaps, while our method doesn't require trimaps. For a fair comparison, we evaluate the whole image for all comparing methods.  Besides, Conn metric fails on some results, and some methods do not estimate the foreground image. We use ``-'' to denote them. 

\subsection{Comparison with state-of-the-art methods}

We compare our methods with the state-of-the-art trimap-based methods, including Index Matting~\cite{lu2019indices}, Context Matting ~\cite{hou2019context}, GCA Matting~\cite{li2020natural} and FBA Matting~\cite{forte2020f}. We also compare with the state-of-the-art background-based method BGMV2 ~\cite{BGMv2}. We used the code / model shared by their authors.  

Table~\ref{table:comp_video240} reports the quantitative results on the Video240K SD dataset. For each trimap-based method, we report their results on the trimap $\mathbf{T}_{seg}$ which is generated from the semantic segmentation, and $\mathbf{T}_{gt}$ which is generated from the ground truth alpha maps. For the background-based method, we assume that users can capture the background image for the first frame and adopt it as the background for the whole video, whose result is denoted as $\mathbf{B}_{first}$. We also report the results with the ground truth background for each frame, which is denoted as $\mathbf{B}_{gt}$. As shown in Table~\ref{table:comp_video240}, our method outperforms both trimap-based and background-based approaches by a large margin. Specifically, our method wins 0.58 on SAD, 0.60 on MSE, 1.46 on Grad, and 0.65 on Conn for the alpha maps. Our method wins 1.37 on MSE for the foreground. Figure~\ref{fig:comp_video240k} shows several visual examples on the Video240k SD dataset. Our results contain more fine structures. In the first example, the girl's hair is our result is much more accurate than others. In the second example, the stethoscope and arms in our result preserve more details.

\subsection{Ablation study}
In this section, we examine several key components of our method.

\begin{table}
    \centering
    \small
    \begin{tabular}{P{1.8cm}P{0.62cm}P{0.62cm}P{0.62cm}P{0.62cm}P{1.25cm}}
            \toprule
            \multirow{2}[2]{*}{Method} & \multicolumn{4}{c}{Alpha} & Foreground  
            \\ \cmidrule(lr){2-5} 
              &SAD &MSE &Grad &Conn &MSE
            \\ \midrule
            MobileNetV2 & 0.63 & 0.54 & 0.80 & 0.41 &1.32 \\
            ResNet18 & 0.48 & 0.27 &0.48 & 0.29 & 1.51 \\ 
            ResNet50 & \textbf{0.44} & \textbf{0.27} & \textbf{0.38} & \textbf{0.26} & \textbf{0.88}  
            \\ \bottomrule
    \end{tabular}\vspace{-0.1in}
        \caption{The effectiveness of the feature extraction network on the Video240K SD dataset.\label{table:comp_ext}} \vspace{-0.1in}
\end{table}

\textbf{Feature extraction network}. We examine how the feature extraction network affects the final results. In this experiment, we change the feature extraction network while fixing the other parts of the network. Table~\ref{table:comp_ext} reports the results of MobileNetV2~\cite{mobilenetv22018}, ResNet18~\cite{he2016deep} and ResNet50~\cite{he2016deep}. Compared with other backbones, ResNet50 is deeper and has higher computational complexity. It achieves the best performance. We believe that these improvements come from the more representative information of ResNet50, which is consistent with the finding of the previous semantic segmentation methods ~\cite{chen2017rethinking,deeplabv3plus2018} and image matting methods~\cite{forte2020f}. 

\begin{table}
    \centering
    \small
    \begin{tabular}{P{1.8cm}P{0.62cm}P{0.62cm}P{0.62cm}P{0.62cm}P{1.25cm}}
            \toprule
            \multirow{2}[2]{*}{Method} & \multicolumn{4}{c}{Alpha} & Foreground  
            \\ \cmidrule(lr){2-5} 
              &SAD &MSE &Grad &Conn &MSE
            \\ \midrule
            MaskFlownet &0.73 & 0.82 & 0.94 & 0.52 &1.61 \\
            PWC-Net &\textbf{0.63} & \textbf{0.54} & \textbf{0.80} & \textbf{0.41} & \textbf{1.32}
            \\ \bottomrule
    \end{tabular}\vspace{-0.1in}
        \caption{The effectiveness of the optical flow network on the Video240K SD dataset.\label{table:comp_flow}} \vspace{-0.2in}
\end{table}

\textbf{Optical flow network}. We examine how the optical flow network affects the final results. In this experiment, we use the MobileNetV2~\cite{mobilenetv22018} as our feature extraction network. We compare two models: one with PWC-Net~\cite{sun2018pwc} as the optical flow network and the other one with MaskFlownet~\cite{zhao2020maskflownet} as the optical flow network. Although MaskFlownet achieves better performance in the optical flow estimation task~\cite{zhao2020maskflownet}, PWC-Net gets better matting results. As shown in Table~\ref{table:comp_flow},  PWC-Net wins 0.10 on SAD, 0.28 on MSE, 0.14 on Grad, and 0.09 on Conn on the alpha maps. It also wins 0.29 on the foreground MSE. This is consistent with many other works that PWC-Net has great generalization capability ~\cite{niklaus2018context,Niklaus_CVPR_2020,bao2019depth}.

\textbf{Context motion updating operator}. We also investigate the contribution of each component in the context motion updating operator. In this experiment, we adopt the MobileNetV2~\cite{mobilenetv22018} as the feature extraction and PWC-Net~\cite{sun2018pwc} as the optical flow network. In particular, we study the impact of context motion updating operator by removing it from the whole network. For the model with ``Baseline'', we remove the context motion updating operator. For the model with ``+ConvGRU'', we only keep the ConvGRU while removing the other components from the context motion updating operator. For the model with ``+Motion'', we keep the context motion updating operator. The performance of these models can be found in Table~\ref{table:comp_cmu}. We can find that ConvGRU can improve all the evaluation metrics on the alpha maps while slightly increasing the foreground MSE. Particularly, the motion plays a crucial role, which improves the results by 0.20 on SAD, 0.42 on MSE, 0.25 on Grad, and 0.16 on Conn. It also improves the foreground MSE by 0.51. It verifies our motivation to introduce the motion information into the video matting network.

\begin{table}
    \centering
    \small
    \begin{tabular}{P{1.8cm}P{0.62cm}P{0.62cm}P{0.62cm}P{0.62cm}P{1.25cm}}
            \toprule
            \multirow{2}[2]{*}{Method} & \multicolumn{4}{c}{Alpha} & Foreground  
            \\ \cmidrule(lr){2-5} 
              &SAD &MSE &Grad &Conn &MSE
            \\ \midrule
            Baseline & 0.95 & 1.45 & 1.22 & 0.75 & 1.66 \\
            + ConvGRU  & 0.83 & 0.96 & 1.05 & 0.57 & 1.83 \\
            + Motion &\textbf{0.63} & \textbf{0.54} & \textbf{0.80} & \textbf{0.41} & \textbf{1.32}
            \\ \bottomrule
    \end{tabular}\vspace{-0.12in}
        \caption{The effectiveness of the context motion updating operator on the Video240K SD dataset. We add the components cumulatively from top to bottom. \label{table:comp_cmu}} \vspace{-0.25in}
\end{table}

\section{Limitation and Future Work}
\label{sec:limit}
There are several limitations and drawbacks of the proposed automatic portrait video matting method.

\textbf{Feature extraction}. Our method employs ResNet50 as our feature extraction network to capture semantic information. It will take too much GPU memory on high-resolution videos, e.g., videos of 4k resolution.  In training, the cropping size of training examples on the high-resolution videos needs to be large to capture the semantic information. Besides, the optical flow network also requires large memory. The high computational and memory requirements impede our training on high-resolution videos.

\textbf{Optical flow}. Our method currently relies on PWC-Net~\cite{sun2018pwc} to estimate the optical flow between frames. However, PWC-Net might not produce reliable optical flow for the challenging scenarios, e.g., videos with large motion. In these videos, our method will fail to produce correct matting results. An interesting future direction is to design an optical flow network specifically for video matting.

\section{Conclusion}
This paper presented an automatic portrait video matting method. We designed a context motion network for this task. Our network takes a sequence of frames as inputs and does not require extra inputs. Our network leverages context information and temporal information. Our method first extracts the context features for each frame. It also extracts the optical flow between consecutive frames. This is followed by a context motion updating operator to fuse the context and motion features recurrently. Our experiments showed that our method is able to generate high-quality alpha maps and foreground images. Our experiments also showed that the context motion updating operator is helpful to generate high-quality results.

\bibliography{aaai22} 

\end{document}